# A Bayesian Network Scoring Metric That Is Based on Globally Uniform Parameter Priors

**Mehmet Kayaalp**
Center for Biomedical Informatics
Intelligent Systems Program
University of Pittsburgh
Pittsburgh, PA 15213
*kayaalp@acm.org*

**Gregory F. Cooper**
Center for Biomedical Informatics
Intelligent Systems Program
University of Pittsburgh
Pittsburgh, PA 15213
*gfc@cbmi.upmc.edu*

**Abstract**

We introduce a new Bayesian network (BN) scoring metric called the *Global Uniform* (GU) metric. This metric is based on a particular type of default parameter prior. Such priors may be useful when a BN developer is not willing or able to specify domain-specific parameter priors. The GU parameter prior specifies that every prior joint probability distribution $P$ consistent with a BN structure $S$ is considered to be equally likely. Distribution $P$ is consistent with $S$ if $P$ includes just the set of independence relations defined by $S$.

We show that the GU metric addresses some undesirable behavior of the BDeu and K2 Bayesian network scoring metrics, which also use particular forms of default parameter priors. A closed form formula for computing GU for special classes of BNs is derived. Efficiently computing GU for an arbitrary BN remains an open problem.

## 1 INTRODUCTION

There is a small number of BN scoring metrics that are frequently applied to learn Bayesian Networks (BNs) from observational data. In this paper, we introduce a new BN scoring metric called the *Global Uniform* (GU) metric. We analyze it along with two other frequently used metrics BDeu and K2, both of which use particular types of default parameter priors, as does GU. The GU metric uses uniform priors, and may be useful when a BN developer is not willing or able to specify domain-specific parameter priors.

In this paper, we provide comparative analyses of the BDeu, K2, and GU metrics on a set of examples. Due to space limitations, we cannot analyze both parameter and structure learning. In this report, we focus solely on structure learning.

Section 2 provides a summary of necessary background information on BNs and parameter priors. In Section 3, we introduce the GU metric. Section 4 provides analyses and experimental results on 10 different independent bivariate distributions of binomial random variables. We also compare BDeu, K2, and GU metrics empirically on the pairs of nodes from the Alarm BN. In the last section, we provide further discussion and conclusions about BDeu, K2, and GU metrics.

## 2 BACKGROUND

A Bayesian network $B = (S, \theta)$ consists of a graphical model $S$ (structure) and a set of probabilities $\theta$ (parameters) defined on $S$. The structure $S$ is represented as a directed acyclic graph, in which each node denotes a random variable, and each arc denotes dependence between variables. Finding a BN that fits the data better than other possible BNs requires a search over the model space. Each step of the search involves using a metric in the evaluation of the model. Searching for the most likely network is called "model selection." The success of model selection depends on the efficiency and effectiveness of the search heuristic (assuming the model space is too large to search exhaustively) and on the scoring metric.

Although there are many scoring metrics, such as information theoretic ones (*e.g.*, AIC, BIC, Kullback-Leibler divergence) and conventional goodness-of-fit metrics (*e.g.*, chi-square statistic, Pearson's chi-square statistic, likelihood ratio statistic), our focus is on Bayesian scoring metrics that are used to score BNs. Usage of these model scoring metrics is not limited to model selection only.

## 2.1 BAYESIAN NETWORK SCORING METRICS

The most distinguishing property of Bayesian metrics is the combination of data with subjective prior probabilities on model parameters and model structures, which are denoted herewith as "parameter priors" and "structure priors," respectively. A BN with $n$ variables $\{X_1,...,X_n\}$ can be scored using the metric in Equation (1). That metric assumes parameter independence, parameter modularity, exchangeable data, and Dirichlet prior probabilities, as discussed in (Cooper and Herskovits 1992; Heckerman, Geiger and Chickering 1995).

$$P(S|D) \propto P(S) \prod_{i=1}^{n} \prod_{j=1}^{q_i} \frac{\Gamma(\alpha_{ij})}{\Gamma(\alpha_{ij}+N_{ij})} \prod_{k=1}^{r_i} \frac{\Gamma(\alpha_{ijk}+N_{ijk})}{\Gamma(\alpha_{ijk})}. \quad (1)$$

The score $P(S|D)$ indicates the probability of a BN structure $S$ for a given database $D$. $P(S)$ is a structure prior determined by the network developer; $r_i$ is the finite number of distinct states that variable $X_i = \{x_1,...,x_{r_i}\}$ can take; $q_i$ is the finite number of distinct state combinations of the variables $\mathbf{Pa}_i = \{\mathbf{pa}_1,...,\mathbf{pa}_{q_i}\}$ that are parents of $X_i$ in $S$. The terms $\alpha_{ijk}$ and $N_{ijk}$ indicate the parameter prior and the number of observations, respectively, where $X_i = x_k$, $\mathbf{Pa}_i = \mathbf{pa}_j$, $N_{ij} = \sum_{k=1}^{r_i} N_{ijk}$ and $\alpha_{ij} = \sum_{k=1}^{r_i} \alpha_{ijk}$.

## 2.2 CHOICE OF PRIORS

Ideally, a domain expert, if available, would provide informative priors for structures and parameters, and the search would start with a structure that the domain expert considers the most likely. Unfortunately, using background knowledge is not always feasible due to the prohibitive cost of expertise and the unavailability of detailed background knowledge. In such situations, which are not uncommon, model developers typically use noninformative priors.

Noninformative structure priors typically are based on the following straightforward assumption: For a given database, every structure is equally likely. This assumption is usually at odds with what an expert would assess. Nevertheless, noninformative structure priors are not biased for a particular structure; thus, they do not systematically misguide the model selection process.

Selecting a set of parameter priors, however, is a more difficult task. When one does not have background knowledge about the nature of a domain, a logical choice is to use a noninformative prior so that the data dominate the posterior probability (Box and Tiao 1973). Uniform priors were first suggested by Bayes and are frequently used for this purpose. Using uniform priors has caused some controversy because of undesired properties, such as not always being invariant under one-to-one transformations. Jeffreys' priors, which *are* invariant under one-to-one transformation, are also controversial, primarily for multiparameter models where they can yield undesirable results (Gelman, Carlin, Stern and Rubin 1995, p. 54). Uniform priors are improper if the range of continuous parameter space is unrestricted, which is not the case in the BDeu, K2, and GU metrics.

Unlike Jeffreys' prior, the BDeu, K2, and GU priors, which are defined in Sections 2.3 and 3, are not invariant under parameter transformation. Typically, these latter three priors are applied on parameters that are natural for the domain being modeled. Each of the three embodies a different notion of indifference over a parameter space that has domain-specific meaning. If one of these notions of indifference is appealing relative to a natural set of domain parameters, then arguably invariance is a less critical criterion.

## 2.3 PARAMETER PRIORS USED IN SCORING METRICS

Cooper and Herskovits (1991;1992) introduced the Bayesian scoring metric K2, which is given by Equation (1) when $\forall ijk \; \alpha_{ijk} = 1$. They also proposed an extension to the K2 that uses Dirichlet priors. Heckerman, Geiger and Chickering (1995) showed that if $\forall i$ the sum of the Dirichlet priors $\alpha_0 = \sum_{jk} \alpha_{ijk}$ is constant, then Equation (1) yields the same score for any two Markov equivalent structures given $D$ and a prior network from which the priors are derived. Due to this property (a.k.a. *likelihood equivalence*), they called this metric likelihood-equivalent Bayesian Dirichlet scoring, in short BDe. A special case of BDe called BDeu uses uniform priors where $\alpha_{ijk} = \alpha_0 / (q_i r_i)$, as proposed by Buntine (1991).

## 3 THE GLOBAL UNIFORM METRIC

The Global Uniform (GU) is a BN scoring metric using uniform priors parameterized over the joint probability distribution. The GU prior can be naturally defined as follows for *any* probabilistic representation, not just for Bayesian networks: Any admissible joint state of the model parameters is believed (in the Bayesian sense) to be equiprobable to any other admissible joint state of those parameters, where admissibility is determined by the constraints imposed by the model structure. In other words, the probability distribution over the admissible states of joint parameters is uniform and its integration over all admissible states yields 1.

The fundamental difference between the GU and other BN scoring metrics is that while priors $\{\alpha_{ijk}\}$ of other BN scoring metrics are defined over the joint probability distributions of *local* structures $\{X_i, \mathbf{Pa}_i\}$, priors of the GU are defined over the joint probability distributions of all variables $\{X_1,...,X_n\}$.

### 3.1 SCORING SATURATED MODELS WITH GU

Suppose the structure $S_1$ is $X \rightarrow Y$ (see Figure 1), where $X = 1,2$ and $Y = 1,2$. Let $\theta_1$, $\theta_2$, $\theta_3$, and $\theta_4$ represent $P(X=1, Y=1)$, $P(X=1, Y=2)$, $P(X=2, Y=1)$, and $P(X=2, Y=2)$, respectively, and let $N_1$, $N_2$, $N_3$, and $N_4$ represent the frequency counts of the respective observations in a given database $D$. Notice that $\theta_4 = 1 - \Sigma_{i=1}^{3}\theta_i$. Assuming a uniform Dirichlet prior $\alpha_1 = \cdots = \alpha_4 = 1$, the marginal likelihood is

$$P(D|S_1)$$

$$= \int_0^1\int_0^1\int_0^1 (\theta_1+\theta_2)^{N_1+N_2}(\theta_3+\theta_4)^{N_3+N_4}(\theta_1/(\theta_1+\theta_2))^{N_1}$$

$$\cdot(\theta_2/(\theta_1+\theta_2))^{N_2}(\theta_3/(\theta_3+\theta_4))^{N_3}(\theta_4/(\theta_3+\theta_4))^{N_4}$$

$$\cdot\mathrm{Dir}(\theta_1,\theta_2,\theta_3\,|\,\alpha_1,\ldots,\alpha_4)d\theta_1d\theta_2d\theta_3 \quad (2)$$

$$= \Gamma(4)\int_0^1\int_0^1\int_0^1 \theta_1^{N_1}\theta_2^{N_2}\theta_3^{N_3}\left(1-\sum_{i=1}^{3}\theta_i\right)^{N_4}d\theta_1d\theta_2d\theta_3$$

$$= \Gamma(4)\frac{\prod_{i=1}^{4}\Gamma(1+N_i)}{\Gamma\left(4+\sum_{i=1}^{4}N_i\right)}$$

In Equation (2), $D$ represents a given sequence of records that constitute a database rather than the sufficient statistic of such a database.

Since BNs are defined in terms of conditional probabilities, it is informative to transform prior joint distributions into the conditional ones, that is:

$$\mathrm{Dir}(\theta_x,\theta_{y|x},\theta_{y|\bar{x}}) = |J|\mathrm{Dir}(\theta_1,\theta_2,\theta_3)$$

$$= |J|\frac{\Gamma(\alpha_0)}{\Gamma(\alpha_1)\cdots\Gamma(\alpha_4)}\theta_1^{\alpha_1-1}\theta_2^{\alpha_2-1}\theta_3^{\alpha_3-1}\left(1-\sum_{i=1}^{3}\theta_i\right)^{\alpha_4-1}, \quad (3)$$

where $|J|$ denotes the absolute value of the Jacobian determinant (DeGroot 1984), $\theta_x$, $\theta_{y|x}$, $\theta_{y|\bar{x}}$ represent conditional parameters $P(X=1)$, $P(Y=1|X=1)$, and $P(Y=1|X=2)$, respectively. Although $\alpha_1 = \cdots = \alpha_4 = 1$ and $\alpha_0 = 4$, their numeric values are not assigned for the clarity of the derivation. The absolute value of the Jacobian determinant turns out to be $|J| = \theta_x(1-\theta_x)$; therefore,

$$\mathrm{Dir}(\theta_x,\theta_{y|x},\theta_{y|\bar{x}})$$

$$= \gamma\theta_x(\theta_x\theta_{y|x})^{\alpha_1-1}(\theta_x\theta_{\bar{y}|x})^{\alpha_2-1}\theta_{\bar{x}}(\theta_{\bar{x}}\theta_{y|\bar{x}})^{\alpha_3-1}(\theta_{\bar{x}}\theta_{\bar{y}|\bar{x}})^{\alpha_4-1} \quad (4)$$

$$= \gamma\theta_x^{\alpha_1+\alpha_2-1}\theta_{\bar{x}}^{\alpha_3+\alpha_4-1}\theta_{y|x}^{\alpha_1-1}\theta_{\bar{y}|x}^{\alpha_2-1}\theta_{y|\bar{x}}^{\alpha_3-1}\theta_{\bar{y}|\bar{x}}^{\alpha_4-1}$$

where $\gamma = \Gamma(\alpha_0)/\prod_{i=1}^{4}\Gamma(\alpha_i)$.

Now, we can derive the factorized form of GU for the given structure $S_1$.

$$P(D|S_1) = \gamma\int_0^1\int_0^1\int_0^1 \theta_x^{N_1+N_2}\theta_{\bar{x}}^{N_3+N_4}\theta_{y|x}^{N_1}\theta_{\bar{y}|x}^{N_2}\theta_{y|\bar{x}}^{N_3}\theta_{\bar{y}|\bar{x}}^{N_4}$$

$$\cdot\theta_x^{\alpha_1+\alpha_2-1}\theta_{\bar{x}}^{\alpha_3+\alpha_4-1}\theta_{y|x}^{\alpha_1-1}\theta_{\bar{y}|x}^{\alpha_2-1}\theta_{y|\bar{x}}^{\alpha_3-1}\theta_{\bar{y}|\bar{x}}^{\alpha_4-1}d\theta_xd\theta_{y|x}d\theta_{y|\bar{x}}$$

$$= \gamma\int_0^1 \theta_x^{N_1+N_2+\alpha_1+\alpha_2-1}(1-\theta_x)^{N_3+N_4+\alpha_3+\alpha_4-1}d\theta_x \quad (5)$$

$$\cdot\int_0^1 \theta_{y|x}^{N_1+\alpha_1-1}(1-\theta_{y|x})^{N_2+\alpha_2-1}d\theta_{y|x}$$

$$\cdot\int_0^1 \theta_{y|\bar{x}}^{N_3+\alpha_3-1}(1-\theta_{y|\bar{x}})^{N_4+\alpha_4-1}d\theta_{y|\bar{x}}$$

Notice that each integral in Equation (5) is a beta function; therefore, Equation (5) is equal to Equation (2) when we set $\alpha_1 = \cdots = \alpha_4 = 1$ as in Equation (2).

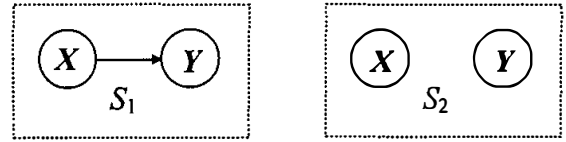

Figure 1: Bayesian networks $S_1$ and $S_2$ each with two binomial variables.

### 3.2 COMBINING GU SCORES OF INDEPENDENT SATURATED MODELS

Consider the BN $S_2$ in Figure 1. Let $N_1$ and $N_2$ represent the number of observations in which $X = 1$ and $X = 2$, respectively, and let $\theta_1 = P(X=1)$. Assuming uniform Dirichlet priors $\alpha_1 = \alpha_2 = 1$, we can obtain the GU score of the binomial variable $X$.

$$P(D|X) = \int_0^1 \theta_1^{N_1}(1-\theta_1)^{N_2}\frac{\Gamma(\alpha_0)}{\Gamma(\alpha_1)\Gamma(\alpha_2)}\theta_1^{\alpha_1-1}(1-\theta_1)^{\alpha_2-1}d\theta_1$$

$$= \Gamma(2)\frac{\prod_{i=1}^{2}\Gamma(1+N_i)}{\Gamma\left(2+\sum_{i=1}^{2}N_i\right)} \quad (6)$$

The GU score of the binomial variable $Y$ is computed analogously. Notice that Equation (6) is equivalent to the BDeu score where $\alpha_0 = 2$.

To compute the GU score of $S_2$, we need to combine the GU scores of marginally independent random variables $X$ and $Y$.

**Lemma.** Let $\theta_{xy} = \{0 \le \theta_{x_i,j} \le 1 \,|\, i = 1,2 \wedge j = 1,2\}$ be the parameter set of the probability distribution $P(X=i, Y=j)$, where $X$ and $Y$ are independent random binomial variables with parameters $0 \le \theta_x \le 1$, and $0 \le \theta_y \le 1$ of the distributions $P(X)$, and $P(Y)$, respectively. If the prior probability distribution $f(\theta_{xy})$ is uniform, then $\forall \theta_x, \theta_y : f(\theta_{xy}) = f(\theta_x)f(\theta_y)$.

**Proof.** Given $\forall \theta_x, \theta_y : f(\theta_{xy})$ is uniform, $\forall \theta_x, \theta_y : f(\theta_{xy}|\theta_y) = f(\theta_{xy}|\theta_x)$, and $f(\theta_{xy})$ defines a plane on a three-dimensional Euclidean space with axes $\theta_x$, $\theta_y$, and $f(\theta_{xy})$, where $\forall \theta_x, \theta_y : f(\theta_{xy}) = 1$. $\forall \theta_x, \theta_y : f(\theta_x) = f(\theta_{xy}|\theta_y) = 1$ and $f(\theta_y) = f(\theta_{xy}|\theta_x) = 1$; thus, $\forall \theta_x, \theta_y : f(\theta_{xy}) = f(\theta_x)f(\theta_y)$. □

The GU score of $S_2$ is equal to the product of the GU scores of the marginally independent variables in $S_2$:

$$P(D|X \perp Y) = P(D|X)P(D|Y)$$

$$= \prod_{i\in\{X,Y\}}\frac{\Gamma(N_{i=1}+1)\Gamma(N_{i=2}+1)}{\Gamma(N_{i=1}+N_{i=2}+2)} \quad (7)$$

Any structure $S$ whose graph representation is a clique (i.e., a saturated directed acyclic graph) with $n$ parameters can be computed using the GU scoring metric as

$$P(D|S) = \Gamma(n)\frac{\prod_{i=1}^{n}\Gamma(N_i+1)}{\Gamma\left(n+\sum_{i=1}^{n}N_i\right)}. \quad (8)$$

When we compare the scores obtained by the GU metric and the BDeu metric on these two examples (dependent and independent binomial bivariate Bayesian networks as shown in Figure 1), we observe that the GU score on the dependent model is equivalent to the BDeu score where $\alpha_0 = 4$ and the GU score on the independent model is equivalent to the BDeu score where $\alpha_0 = 2$. In general, in order to be uniform over all admissible prior joint probability distributions, as is the GU metric, we cannot use some *constant* $\alpha_0$ for all the structures being scored. It follows then that the BDeu metric favors some admissible prior joint probability distributions over others.

### 3.3 LIKELIHOOD EQUIVALENCE

Let $Q$ denote the set of all joint probability distributions that a model $M$ can represent. Each element $\theta$ of $Q$ is a joint distribution that is representable by model $M$. If $\theta$ is in $Q$ we will say that $\theta$ is *consistent with M*. The GU prior for model $M$ is a parameter prior in which $\forall \theta \in Q : f(\theta) = c$ and $\forall \theta \notin Q : f(\theta) = 0$, where $f(\theta)$ denotes the prior probability distribution of $\theta$ and $c$ is a constant.

In this paper, we concentrate on models that are BNs with multinomial likelihood functions, which we call simply *multinomial BNs*. The *GU metric* for scoring such models is defined to be the marginal likelihood that results from using the GU prior. We next show that the GU metric exhibits the likelihood equivalence property (Heckerman et al. 1995).

**Theorem.** Consider two Bayesian network structures $S_1$ and $S_2$ that (1) contain the same set of discrete variables $\mathbf{X} = \{X_1, \ldots, X_n\}$, (2) have parameters that are represented with multinomial distributions, and (3) are independence (Markov) equivalent. Then, using the GU prior will render $S_1$ and $S_2$ likelihood equivalent, that is, for every dataset $D = \{\mathbf{x}\}$: $P(D | S_1) = P(D | S_2)$.

**Proof**: It has been shown that conditions 1, 2, and 3 imply that any joint probability distribution that can be represented by $S_1$ can also be represented by $S_2$, and vice versa (Heckerman et al. 1995). Let $\theta_1$ and $\theta_2$ be the factorizations of $\theta$ according to $S_1$ and $S_2$, respectively. By the definition of the GU metric, for every $\theta$ that can be represented by $S_1$ and $S_2$ we have that: $f(\theta|S_1) = f(\theta_1|S_1) = c$, $f(\theta|S_2) = f(\theta_2|S_2) = c$, and thus, $f(\theta_1|S_1) = f(\theta_2|S_2)$. Likewise, for every $\theta$ that cannot be represented by $S_1$ and $S_2$ we have that: $f(\theta|S_1) = f(\theta_1|S_1) = 0$, $f(\theta|S_2) = f(\theta_2|S_2) = 0$, thus, with a GU prior,

$$\forall \theta : f(\theta | S_1) = f(\theta | S_2). \quad (9)$$

Since $\theta = \{P(\mathbf{x} \in \mathbf{X})\}$, $\forall \mathbf{x} : P(\mathbf{x} | S_1, \theta_1) = P(\mathbf{x} | S_2, \theta_2)$. Since $D = \{\mathbf{x}\}$, it follows from the three conditions in the theorem that:

$$\prod_{\mathbf{x} \in D} P(\mathbf{x} | S_1, \theta_1) = \prod_{\mathbf{x} \in D} P(\mathbf{x} | S_2, \theta_2), \text{ and thus} \quad (10)$$

$$P(D | S_1, \theta_1) = P(D | S_2, \theta_2).$$

Integrating over the product of terms in Equations (9) and (10) we obtain:

$$\int_\theta P(D|S_1, \theta_1) f(\theta|S_1) d\theta = \int_\theta P(D|S_2, \theta_2) f(\theta|S_2) d\theta \quad (11)$$

Thus,

$$P(D | S_1) = P(D | S_2) \quad (12)$$

□

By the definition of GU, $f(\theta | S_1) = f(\theta | S_2) = c$, where $S_1$ and $S_2$ are Markov equivalent BN structures, thus GU is a distribution equivalent metric (Heckerman and Geiger 1995). It is important to note that since the GU metric is defined over the parameter space of joint probabilities, so that neither parameter independence nor parameter modularity holds. For instance, the prior parameterizations of the random variable $X$ in $S_1$ (see Figure 1) and $S_2$ were different—in the former, $\alpha_0 = 4$; whereas, in the latter $\alpha_0 = 2$. Since $P(D | X)$ differ in two different structures depending on the presence of an arc from $X$ to $Y$, parameter modularity does not hold.

## 4 RESULTS

In this section, we analyze GU, BDeu, and K2 metrics on synthetic examples and provide experimental results about their behavior. Each example consists of a different joint probability distribution on two marginally independent variables. In each analysis, we first provide the data-generating distribution, followed by samples of various sizes that mirror the data generating distribution almost exactly; that is, the statistic realized in each dataset were generated by multiplying the underlying joint density distributions by the sample size and rounding each result to its nearest integer value in order to eliminate sampling noise.

### 4.1 TWO OR MORE CONSTANTS IN A DATABASE

In this subsection, we analyze an example that contains two binomial variables $X$ and $Y$, whose marginal probabilities are $P(X = 1) = 1$ and $P(Y = 1) = 1$. These variables are marginally independent, since $\forall X, Y$

$$P(X, Y) = P(X) P(Y). \quad (13)$$

The Dirichlet assumption requires that each hyperparameter be *strictly* positive. Due to this assumption, as well as to $\Gamma(0) = \infty$, we cannot assign zero to any $\alpha_{ijk}$ in Equation (1). If all parameter priors are strictly positive, all posterior probabilities would also be strictly positive. The BDe metric may be considered undefined if the parameter space is not strictly positive, since the derivation of the BDe metric involves the use of Dirichlet distributions that themselves require a strictly positive parameter space (Heckerman et al. 1995). Certainly, it is possible to have data generating distributions, for which $P(X = 0, Y = 0) = 0$.

Consider a database $D$, in which for all cases $d \in D$, $X = 1$ and $Y = 1$. Scores of two structures $S_1: X \rightarrow Y$ and $S_2: X \perp Y$ (see Figure 1) are computed based on BDeu, K2 and GU metrics. In order to observe the relation between the BDeu score and the prior equivalent sample size, three different prior equivalent sample sizes are used, $\alpha_0 = 0.01, 1, 4$, corresponding to BDeu0, BDeu1, and BDeu4, respectively. The results are obtained for three databases with sample sizes 10, 1000, and $10^5$ and plotted in Figure 2, where the y-axis labeled as the "BDeu, K2, and GU ratios" specifies $P(S_1|D)/P(S_2|D)$ as a function of the sample size, which is shown on the x-axis.

In Example 1, the BDeu ratios are always greater than 1, and $\lim \alpha_0 \to 0: P(S_1|D)/P(S_2|D) = 1^+$. In other words, the BDeu favors the dependent structure, even though in the underlying distribution the two variables are marginally independent. This pattern gets stronger as the sample size increases.

The BDeu score for the variable $X$ is the same in both structures. Suppose all variables are binomial (as in our examples); then, in Equation (1) the following equations hold: $q_Y^{S_2} = q_Y^{S_1}/2$ and $\alpha_{Y_{jk}}^{S_2} = \alpha_{Y_{jk}}^{S_1}/2$. Since both variables are constant, there is only one configuration observed; i.e., for both structures, there exists a unique $Y_{jk}$ for which $N_{Y_{jk}}$ is equal to sample size $N$, and for all other configurations $N_{Y_{jk}} = 0$. After substitutions and cancellations, the BDeu ratio for Example 1 is

$$P(S_1|D)/P(S_2|D)$$
$$= \frac{\left(\Gamma(\alpha_0/q_i^{S_1})\right)^2 \Gamma(N+\alpha_0/2q_i^{S_1})\Gamma(N+2\alpha_0/q_i^{S_1})}{\Gamma(\alpha_0/2q_i^{S_1})\Gamma(2\alpha_0/q_i^{S_1})\left(\Gamma(N+\alpha_0/q_i^{S_1})\right)^2} \quad (14)$$
$$> 1$$

Given $N > 1$, the terms containing $N$ in Equation (14) dominate the score, and since its numerator is always greater than its denominator, the score is greater than 1 for every strictly positive $\alpha_0$ and $q_i$. The ratio gets larger for the larger sample size $N$ and/or for the larger equivalent sample size $\alpha_0$.

In the general case, given that there are $n$ variables that are observed constant in a given finite sample $D$, the BDeu metric gives the highest score to each structure that consists of a clique of these $n$ marginally independent variables. The proof involves an inductive extension of the above proof for the binomial bivariate case. As the number of such variables increases and/or the prior sample size decreases (as in BDeu0), the slopes in Figure 2 decrease, but remain positive.

This behavior of the BDeu may sometimes cause problems in high dimensional domains (such as multivariate nonstationary time-series) with sparse data, since the metric may yield complex local structures that consist of unrelated variables when the database contains only single (constant) instantiations of such variables. We encountered this problem in our own research in using BDeu to construct dynamic Bayesian networks from medical data; that experience prompted us to pursue the research reported here. To our knowledge, this problem has not been reported in the literature.

In this example, K2 remains at 1 for all sample sizes. Thus, K2 takes the lack of variation as providing no information about dependence among $X$ and $Y$. The GU metric favors the independent structure more strongly as the sample size grows. Thus, GU takes the lack of variation as indicating that the two variables are independent. In fact, they are marginally independent. For this example scenario, GU arguably provides the most appropriate scores, at least in the large sample limit.

### 4.2 SKEWED MARGINAL DISTRIBUTIONS

The following four examples contain two independent binomial variables. In Examples 2–5, $P(Y=1) = 0.999$, and $P(X=1)$ varies among the values 0.999, 0.9, 0.7, and 0.5, respectively. Databases of size 1000 are generated to mirror their joint probabilities without adding any noise. In Example 2, for instance, the number of cases for $(X,Y) = \{(1,1),(1,2),(2,1),(2,2)\}$ are 998, 1, 1, and 0, respectively.

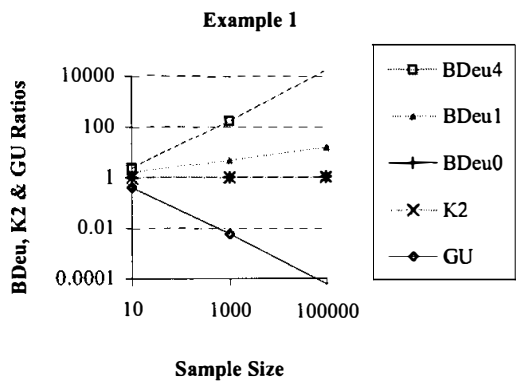

Figure 2: Example 1 contains two marginally independent variables $X \perp Y$ that have constant values in the database. BDeu favors for the wrong structure, K2 stays indiscriminative and GU favors for the independent structure.

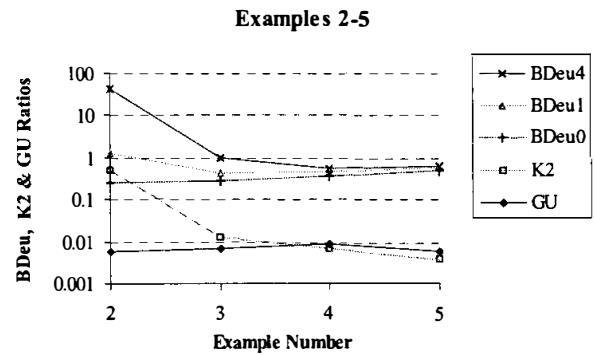

Figure 3: In the generating model, $P(Y=1) = 0.999$; whereas, $P(X=1)$ changes from 0.999 to 0.9, 0.7, and 0.5 in Examples 2–5, respectively.

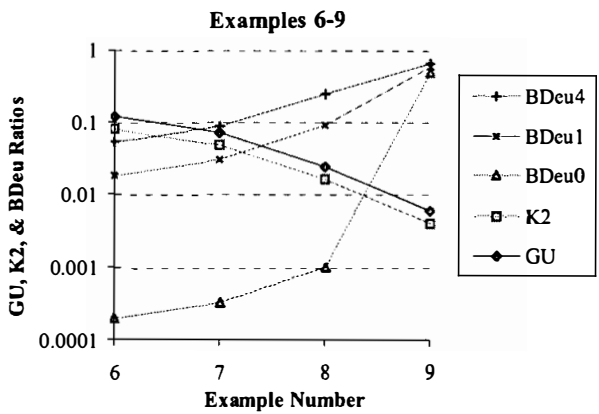
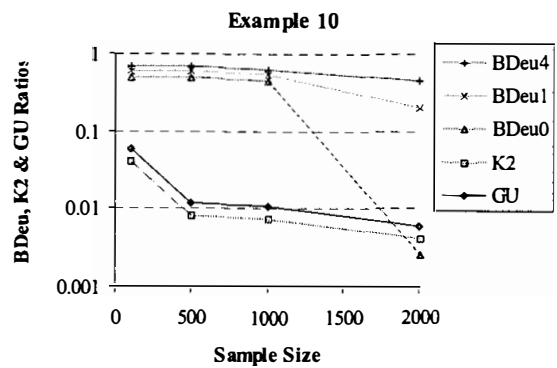

Figure 4: In Examples 6–9, $P(Y=1)=0.5$, whereas $P(X=1)$ varies between 0.5, 0.9, 0.99, and 0.9995, respectively.

Figure 5: Effect of sample size on BDeu, K2, and GU metrics, where $P(X=1)=0.999$ and $P(Y=1)=0.55$.

Given $\alpha_0 = 4$, the BDeu metric favors the dependent structure in Examples 2 and 3, but not in Examples 4 and 5 (see Figure 3). Ratios of the BDeu-based model scores change significantly, depending on the degree of the skewness of the distribution of a variable and the prior sample size. One might expect that when $\alpha_0$ increases, the BDeu metric would increasingly favor variable independence; however, Examples 2–5 illustrate that this need not occur; i.e., as $\alpha_0$ increases, the BDeu ratios get larger.

GU ratios vary minimally as the marginal distributions of variables change; i.e., the GU metric seems robust in determining the independence relationships between variables and immune against the marginal distributional variations of variable $X$. For skewed distributions, as in Examples 2 and 3, BDeu ratios vary significantly; whereas, they get closer to each other as the data match priors more closely, as in Examples 4 and 5. Throughout Examples 2–5, GU ratios are closer to zero than are the BDeu ratios. The same is true for K2 with respect to BDeu ratios in Examples 3–5.

### 4.3 MATCH BETWEEN PRIOR AND DATA

In Examples 6–9, the sample size is 1000 and the generating structure is $S_2$ with $P(X=1)=0.5$, 0.9, 0.99, and 0.9995, and $P(Y=1)=0.5$. In these examples, the BDeu ratio is always less than 1. Since the data generating structure is $S_2$, the BDeu score would lead to correct model selection. As $P(X=1)$ gets closer to 1, BDeu ratios approach 1 as well.

The GU ratios in Examples 6–9 do not stay as stationary as they do in Examples 2–5, and along with K2 they approach 0 as the skewness of $X$ increases. The rate of change of the BDeu0 ratio is very sharp, indicating the sensitivity of this metric to changes in the generating distribution.

### 4.4 THE EFFECT OF SAMPLE SIZE

In Example 10, four databases of size 100, 500, 1000, and 2000 were generated from two independent binomial variables with distributions $P(X=1)=0.999$ and $P(Y=1)=0.55$, without adding any noise. As expected, all scoring metrics move toward zero as the sample size increases, as shown in Figure 5. For the BDeu ratios, as $\alpha_0$ decreases, the steepness of the slope of the BDeu ratios increases, and data (rather than priors) dominate posteriors more strongly. For sample sizes $N \leq 1000$, GU and K2 ratios are closer to zero relative to the BDeu ratios. The BDeu0 changes that relation at $N = 2000$.

### 4.5 BDeu, $\alpha_0$ AND SAMPLE SIZE

Prior equivalent sample size $\alpha_0$ plays a major role in model selection. In Example 11, we analyze the behavior of the BDeu by varying $\alpha_0$ and sample sizes. The sample sizes and the generating joint distributions are the same as in Example 10. The BDeu ratios are plotted in Figure 6. As seen in these plots, each BDeu ratio reaches a maximum at a certain value of $\alpha_0$; the larger the sample size, the larger the BDeu ratio and the value of $\alpha_0$. For all $\alpha_0 > 250$ and $N > 500$, the BDeu score consistently favors the dependent structure over the independent one and this pattern is stronger in the larger samples. We also found that this maximum depends on the skewness of the distribution. For instance, if $P(Y=1)$ shifts from 0.55 as in Example 11 to 0.9, the value of the maximum BDeu ratio (for $N = 1000$) changes from being 1.9 as plotted in Figure 6 to being $10^{26}$.

### 4.6 AN EMPIRICAL EVALUATION BASED ON DATA GENERATED FROM THE ALARM BAYESIAN NETWORK

We examined the performance of the BDeu, K2, and GU metrics in scoring the probability that there is an arc between pairs of nodes in the Alarm BN (Beinlich, Suermondt, Chavez and Cooper 1989). We considered two sets of pairs. Set $A$ consists of the pairs of nodes for which there actually is an arc in Alarm; there are 46 such pairs. The other set, denoted as $I$, consists of nodes that are d-separate in Alarm; out of the 300 such pairs, we randomly selected 46.

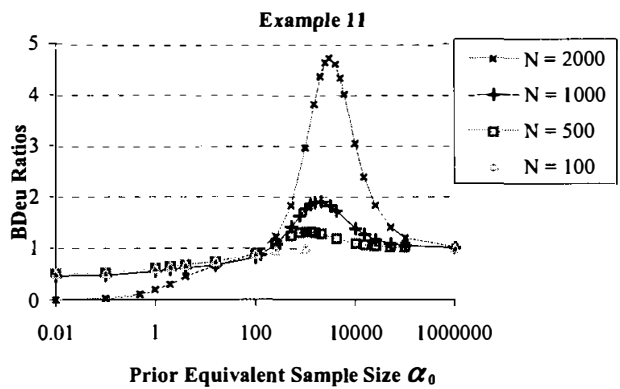

Figure 6: Given a large prior equivalent sample size $\alpha_0$, as the actual sample size grows, the BDeu metric increasingly scores the dependent structure as more probable than the independent structure.

Using BDeu, K2, and GU, we scored each of the arcs in $A$ and in $I$ with datasets of varying sizes. Each dataset was generated by random sampling from the distribution specified by Alarm. We considered datasets that contain 5, 10, 20, 40, 80, and 160 complete cases (no missing data).

We considered three versions of BDeu, corresponding to an $\alpha_0$ of 0.01, 1, and 4. For each dataset of a given size, we applied GU, K2, and the three versions of BDeu to determine the posterior probability of each arc in $A$. From these probabilities we generated an ROC curve for each of the five methods. The ROC curve captures the true-positive rate of arc classification versus the false-positive rate, as the probability classification threshold is varied from 0 to 1.

We generated 100 datasets of a given size. For each dataset, we generated an ROC curve for each of the five methods. Each ROC curve has an area, which serves as one measure of predictive performance. For each method, we computed the area of all 100 ROC curves, then took the mean area as an overall statistic of performance.

The results indicate that BDeu1 and BDeu4 perform quite well in that the areas under their ROC curves (AUC) are larger than the AUC for the GU metric in all datasets. The largest AUC difference we found was between BDeu4 and GU on a dataset containing 20 cases (see Figure 7). The error bars indicate 95% confidence intervals.[1] The mean ROC curves of GU and K2 were within the range of the error bars of the BDeu4, except one ROC point of GU whose sensitivity distance from the error boundary of the BDeu4 was 0.02. BDeu0 performed slightly better than GU only on the 10-case datasets, they were similar in performance on the 20-case datasets, and BDeu0 had a lower AUC on all other datasets.

---

[1] Error bars are generated using a $t$ distribution.

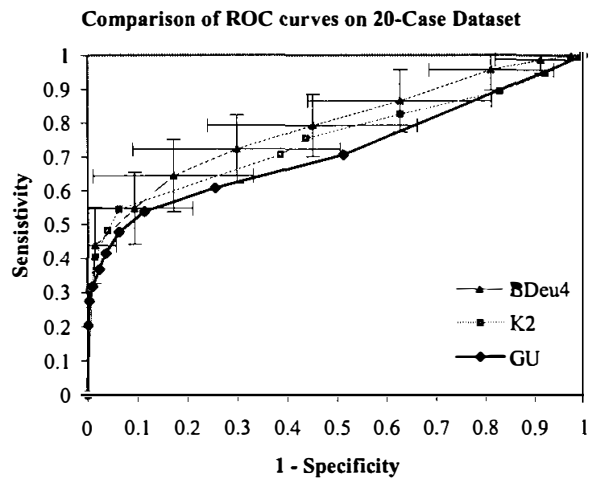

Figure 7: Mean ROC curves of BDeu4, K2 and GU.

## 5 DISCUSSION

In this study, we introduced the GU metric to score BNs when priors based on domain knowledge are unavailable or not used. In Example 1, we analyzed a problem with the BDeu metric, which may be significant in certain domain problems and datasets. Three BN scoring metrics BDeu, K2 and GU were further analyzed on various marginally independent distributions and on pairs of nodes from the Alarm BN.

To obtain a GU score efficiently, we provided a closed form formula for networks that are represented by one or more cliques (i.e., saturated directed acyclic graphs).

The appropriateness of using uniform priors has been an area of discussion among Bayesian statisticians, ever since the concept was introduced by Bayes. There are certain extreme situations in which uniform distributions behave unexpectedly under transformations (Box and Tiao 1973, pp. 23–25). A uniform prior is often found appropriate for statistical inference when it lets the data, rather than the priors, dominate posterior probabilities (Box and Tiao 1973, pp. 21–23). The unexpected behavior of the BDeu scoring metric may be associated with the dominance of its prior over the likelihood. In such situations, when no prior knowledge is available, GU seems to offer a robust alternative to the BDeu metric.

In Examples 1–10, we observed that the BDeu metric tends toward dependency. This tendency is proportional to the skewness of the independent variables, the sample size, and the prior equivalent sample size. Results on the Alarm network data also show that BDeu with $\alpha_0 = 1$ and 4 tends to add arcs more than does GU. In Examples 1–10, GU ratios identify independence between variables more readily than do BDeu ratios.

In Examples 3–10, the behavior of K2 is very similar to that of GU. In Example 1, although K2 does not identify the independence between two marginally independent variables, it does not score them as dependent; rather, its score reflects that the data cannot discriminate the two structures. In Example 2, the K2 metric scores similar to the BDeu metrics that use small prior equivalent sample sizes.

In Example 1, BDeu failed to identify that $X$ and $Y$ are marginally independent, while the GU did identify it. It may be argued, however, that BDe was not designed for situations in which some parameters (marginal distributions of variables in Example 1) are not strictly positive. Such situations occur frequently, however, in certain circumstances, such as nonstationary time series modeling, in which high dimensionality and sparseness of datasets are common issues. Indeed, in our own research, we have encountered such circumstances. When the BDeu metric is used in such situations, we offer the following suggestion: To prevent adding extraneous arcs that provide no predictive power, do not consider as a parent any node that has only one value in the database.

In examples 2–10, in which all parameters are strictly positive, observations on some behavior of the BDeu metric seem counterintuitive: increases of the uniform prior equivalent sample size do not yield stronger marginal independence, even though the data indicate that variables are marginally independent.

The 11 examples introduced in this study provide benchmarks that may be useful in the investigation of new scoring metrics in the future.

Tests with the Alarm network showed that when the distributions are not as skewed as in nonstationary time-series and our synthetic data sets (Examples 1–10), then the assumptions used in the BDeu metric may be desirable. It is interesting however that the BDeu0 results were not as desirable as the results obtained with the other two BDeu metrics.

Based on all the results reported here, our current assessment is as follows. The BDeu metric is likely to perform better with $\alpha_0$ values in the range of 1 to 4, than values near 0. There exists, however, data-generating distributions for which BDeu does not do well in learning the generating structure, even when using $\alpha_0$ in the range of 1 to 4. These situations involve distributions that contain variables with values that are fixed (i.e., only one value is obtained) or close to being fixed. In these situation, special algorithmic checks should be considered when applying BDeu. Finally, empirical results from using one domain network suggest that BDeu performs well in practice. Indeed, it outperformed GU in terms of area under the ROC curve, although not usually statistically significantly so. We are still investigating the reason for this result. One possibility is that parameter independence holds approximately in the Alarm network, and thus, metrics that assume parameter independence (including BDeu and K2) might thereby perform better than those that do not (such as GU).

Future research includes developing an exact or approximate method for computing the GU metric for an arbitrary BN and evaluating the metric on a wider set of synthetic and real databases.

### Acknowledgments

We thank Peter Spirtes and the UAI anonymous reviewers for comments on an earlier version of this paper. This research was supported in part by grant IIS-9812021 from the National Science Foundation and by grants 2T15LM/DE07059-12, G08-LM06625, LM06696 and LM06759 from the National Library of Medicine.

### References

Beinlich, I. A., Suermondt, H. J., Chavez, R. M., and Cooper, G. F. (1989), "The ALARM monitoring system: A case study with two probabilistic inference techniques for belief networks," in *Proceedings of the Second European Conference on Artificial Intelligence in Medicine*, 247–256.

Box, G. E. P. and Tiao, G. (1973), *Bayesian Inference in Statistical Analysis*. London, Addison-Wesley.

Buntine, W. L. (1991), "Theory Refinement on Bayesian Networks," in *Proceedings of the Seventh Conference on Uncertainty in Artificial Intelligence*.

Cooper, G. F. and Herskovits, E. (1991), "A Bayesian Method for Constructing Bayesian Belief Networks from Databases," in *Proceedings of the Seventh Annual Conference on Uncertainty in Artificial Intelligence*.

Cooper, G. F. and Herskovits, E. (1992), "A Bayesian Method for the Induction of Probabilistic Networks from Data," *Machine Learning*, 9, 309–347.

DeGroot, M. H. (1984), *Probability and Statistics*. Reading, Mass., Addison-Wesley.

Gelman, A., Carlin, J. B., Stern, H. S., and Rubin, D. B. (1995), *Bayesian Data Analysis*. London, Chapman & Hall.

Heckerman, D. and Geiger, D. (1995), *Likelihoods and Parameter Priors for Bayesian Networks*. Tech. Rep. MSR-TR-95-54, Microsoft Research.

Heckerman, D., Geiger, D., and Chickering, D. M. (1995), "Learning Bayesian Networks: The Combination of Knowledge and Statistical Data," *Machine Learning*, 20, 197–243.